\documentclass{article}
\usepackage{iclr2027_conference,times}
\usepackage{float}

\usepackage{amsmath,amsfonts,bm}

\def\eqref#1{equation~\ref{#1}}

\def\1{\bm{1}}

\def\eps{{\epsilon}}

\DeclareMathAlphabet{\mathsfit}{\encodingdefault}{\sfdefault}{m}{sl}
\SetMathAlphabet{\mathsfit}{bold}{\encodingdefault}{\sfdefault}{bx}{n}

\usepackage{hyperref}
\usepackage{url}
\usepackage{xurl}
\usepackage{graphicx}
\usepackage{booktabs}
\usepackage{tikz}
\usetikzlibrary{positioning,arrows.meta,decorations.pathreplacing,calc}

\title{Trained Agentic Context Management}

\author{Bryce Sandlund \\
\url{https://brycesandlund.com}
}

\iclrfinalcopy   %
\makeatletter
\def\@maketitle{\vbox{\hsize\textwidth
{\centering\LARGE\sc \@title\par}
\lhead{Preprint}
{\centering\begin{tabular}[t]{c}\bf\rule{\z@}{24pt}\@author\end{tabular}\par}
\vskip 0.3in minus 0.1in}}
\makeatother
\begin{document}

\maketitle

\begin{abstract}
We study long context language models. Instead of training long context natively, or designing a long context harness, we train a model over the simplest possible harness: a tool to call itself with any specified prompt and a tool to read tokens in a range from the input context. We finetune Qwen3.6-35B-A3B on a diverse synthetic dataset using this harness. With only 8,000 tokens of context, our small model is as strong as GPT-5.4 with 1M tokens of context on the OOLONG-synth benchmark when document length exceeds 40K tokens\footnote{Our code is available at: \url{https://github.com/brycesandlund/infinite-context}.}.
\end{abstract}

\begin{figure}[H]
\centering
\begin{minipage}[t]{0.49\linewidth}\centering
\includegraphics[width=\linewidth]{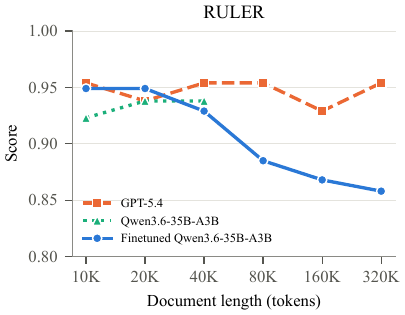}
\end{minipage}\hfill
\begin{minipage}[t]{0.49\linewidth}\centering
\includegraphics[width=\linewidth]{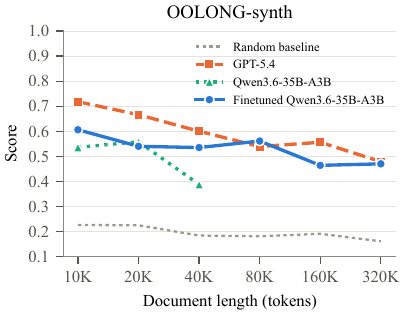}
\end{minipage}
\caption[Results on RULER and OOLONG-synth]{\textbf{Results} on RULER and OOLONG-synth. Our finetune runs in our harness with an 8K token
per-agent context limit. GPT-5.4 (medium reasoning) and base Qwen3.6-35B-A3B (thinking disabled) receive the full document with the prompt (see Section~\ref{experiments} and Appendix~\ref{app:harness} for more details). Base Qwen and GPT-5.4 are served with 64K and 1M token context limits, respectively. The OOLONG random
baseline follows \citet[\S2.3]{bertsch2025oolong}.}
\label{fig:length-sweep}
\end{figure}

\section{Introduction}

As language models boast impressive breakthroughs~\citep{luong2025imogold,wei2025imogold,openai2025icpc,lin2025icpcgold,openai2026unitdistance,openai2026tenadvances,openai2026navierstokes}, increasing focus is placed on their limitations. The most salient resource constraint of a language model is its context length. Models can process and produce tokens within this limit, but the limit is hard. Model producers have identified this constraint and placed great emphasis on extending it as large as possible, with some model builders touting 10M~\citep{dangel2026subq,meta2025llama4,zhu2026pokeeisaac} and even 100M~\citep{magic2024ltm2mini} token context lengths (for reference, the length of the first Harry Potter book is about 100,000 tokens, the full U.S. tax code roughly 3M, and the Linux kernel source code about 300M).

While native context scaling is appealing as a concrete engineering problem, it is not the only approach to producing a model capable of processing long documents. In 2021, OpenAI fine-tuned GPT-3 to summarize sections of books, and then summarize those summaries, achieving competitive performance with human summaries on a model with a context length of just 2,048 tokens~\citep{wu2021recursive}. As the field has matured, agentic harnesses have become viable. MemGPT treats context similarly to an operating system, relying on LLM-invoked function calls to push relevant information in and out of the context window to external memory, achieving effectively expanded virtual context~\citep{packer2023memgpt}. More recently, Recursive Language Models introduced a harness that leverages LLMs' coding prior: the model is repeatedly invoked in a REPL environment over $O(1)$-sized result snippets, achieving stronger performance than base models on retrieval and aggregation tasks both within context limits, but also beyond them~\citep{zhang2025rlm}.

While handcrafted prompts and programmatic environments can be effective in solving long context problems without native long context training, it begs an obvious question: if models can, for example, solve all the problems at the International Collegiate Programming Contest\footnote{\url{https://icpc.global/}} World Finals~\citep{openai2025icpc}, or more generally, implement requested features in codebases far larger than context limits~\citep{marten2026terminalbench,openai2026gpt6astra,anthropic2026fablemythos51}, can they not be \emph{trained} to manage context effectively, \emph{without human-authored harness crutches}? Such model training internalizes decisions external harnesses hardcode, offering the potential to scale much more generally.

We answer this question in this paper, affirmatively. We consider the simplest possible agentic context harness, defined below. The purpose of the harness is to \emph{enable capability}, without \emph{fixing choices} or \emph{instilling a prior}.

\subsection{Harness}

Our harness has two tools:

\begin{enumerate}
  \item \verb|read_chunk(start, end)|: Read a slice of the document and return the decoded text of tokens [\verb|start|, \verb|end|).
  \item \verb|spawn_subagent(subtask)|: Spawn a fresh-context copy of yourself to solve \verb|subtask|; returns the child's \verb|\boxed{}| answer.
\end{enumerate}

Unmodified long-context prompts are provided to the model over a document accessed via the above harness. The system prompt is propagated to child agents and includes the above harness definition, a generic instruction, and task-specific background, in accordance with the respective eval\footnote{Precise harness environment is given in Appendix~\ref{app:harness}.}. Subagents' user prompt is specified by the calling model.

\subsection{Results}

We finetune Qwen3.6-35B-A3B on a diverse synthetic dataset using this harness and achieve the following results (see Figure~\ref{fig:length-sweep}):
\begin{enumerate}
  \item On the OOLONG-synth benchmark, with a context limit of just 8K tokens per agent, our small model performs \textbf{as well} as GPT-5.4 with 1M tokens of context when document length exceeds 40K tokens.
  \item On the RULER benchmark, with a context limit of just 8K tokens per agent, our model scores $\geq85\%$ on document size from 10K to 320K tokens.
\end{enumerate}

\subsection{Paper Organization}

In the next section, we survey related work to frame the contribution of the following sections. In Section~\ref{methods}, we describe our training methodology. In Section~\ref{experiments}, we discuss our experiments, and we conclude in Section~\ref{conclude}.

\section{Related Work}
\label{related}

Long-context LLM work can be broken into a few distinct categories: \emph{native long context models}, \emph{context compaction}, and \emph{long context harnesses}, both trained and untrained.

\subsection{Native Long Context Models}

GPT-2 shipped with a 1,024 token context limit~\citep{radford2019gpt2}. This limit doubled in every generation through GPT-4~\citep{brown2020gpt3,openai2022chatgpt,openai2023gpt4}, until jumping to the 100K range on GPT-4 Turbo~\citep{openai2023devday}, with contemporaneous models Claude 2~\citep{anthropic2023claude2} and later Llama 3.1~\citep{grattafiori2024llama3} at similar limits. The core difficulty in scaling context length $n$ in the classic Transformer architecture~\citep{vaswani2017attention} is $\Omega(n^2)$ attention complexity, which becomes prohibitively expensive as $n \gg 100$K. State-space architectures like the mamba lineage~\citep{gu2023mamba,dao2024mamba2,lahoti2026mamba3}, neural methods like Titans~\citep{behrouz2024titans}, and Linear Transformer~\citep{katharopoulos2020lineartransformers} derivatives DeltaNet~\citep{schlag2021fastweight} and Gated DeltaNet~\citep{yang2025gateddeltanet} abandon quadratic attention for recurrence. Modern architectures are hybrid, combining linear or sparse attention layers with dense attention, allowing context lengths to scale to 1M+ tokens while retaining long-context performance~\citep{qwen2026qwen35,kimiteam2026kimik3,deepseekai2025deepseekv32,glm5team2026glm5,nvidia2025nemotron3,gemmateam2024gemma2,mistral2024ministral}.

\subsection{Context Compaction}

An alternative to native long context, and an inevitability when such native context fills, is a function $f$ mapping current context $x$ to a reduced context $y$ where $|y| \ll |x|$. Three categories of such lossy $f$ exist: \emph{masking}, \emph{summarization}, and \emph{learned compaction}. The simplest compaction technique is masking: dropping tokens based on a predetermined rule. Clearing tool call results is a popular recommended choice~\citep{anthropic2025contextengineering}, achieving results competitive with summarization~\citep{deepseekai2025deepseekv32,lindenbauer2025complexitytrap}. Alternative approaches evict tokens based on attention statistics~\citep{xiao2023streamingllm,zhang2023h2o,li2024snapkv}. Summarization is the current production harness default, implemented in Claude Code, Codex, and Gemini CLI~\citep{anthropic2025contextengineering,openai2026codexrepo,google2026geminicli,barbaste2026harnessengineering}. Trained compaction is a new frontier, described as part of GPT-5.1-Codex-Max~\citep{openai2025codexmax} and Cursor's composer~\citep{cursor2026selfsummarization} training. ``Still'' trains a model to compress KV-Cache directly~\citep{oneill2026still}.

\subsection{Long Context Harnesses}

As long as LLMs have proven useful, an evolving line of work has focused on how to expand their capabilities without touching model internals. The first efforts were known as prompt engineering~\citep{phoenix2024promptengineering,berryman2024promptengineering}, which, though necessary on pre-trained base models~\citep{brown2020gpt3}, became internalized through instruction-tuning and reasoning RL~\citep{ouyang2022instructgpt,guo2025deepseekr1}. Function-calling has led to a new wave of such efforts, where domain-specific knowledge, tools, and multi-agent systems live on a scaffold that interacts with powerful LLMs known as a ``harness"~\citep{lopopolo2026harnessengineering,barbaste2026harnessengineering}. 

Long-context harnesses originate with systems like MemGPT~\citep{packer2023memgpt}, RAPTOR~\citep{sarthi2024raptor}, and Chain of Agents~\citep{zhang2024coa}, where the overall goal is to engineer workflows that decompose documents into contexts independent LLM calls can read and aggregate. Recursive Language Models is a popular recent example that allows for controlled-depth recursive LLM calls via a programmatic environment~\citep{zhang2025rlm}. Follow-up work expands subagent capabilities~\citep{lumer2026rah}, trains shared parent/child policies~\citep{kim2026reinforcingrlm}, and incorporates self-reflection~\citep{alizadeh2026srlm}.

Memagent~\citep{yu2025memagent} trains a harness that walks linearly through context, using RL and a harness prompt to determine what to carry from node-to-node. Similarly to our work, they also use a small (8K token) context while processing long (as much as 3.5M token) documents. However, they test primarily on the distribution they train, compare only against other small models, and hardcode execution path and prompts in the harness.

Frontier models now utilize multi-agent systems. Kimi K2.5 trains ``agent swarm"~\citep{kimiteam2026kimik25}: the model is trained to call subagents via reinforcement learning. Subagent trajectories are not trained. OpenAI has also trained multi-agent behavior~\citep{openai2026gpt56}, notably utilizing 10,000 agents to solve Navier-Stokes~\citep{openai2026navierstokes} and disclosing cooperatively trained agents in the HuggingFace incident~\citep{openai2026hfincident,greenblatt2026hfinvestigation}. Anthropic, DeepMind, Meta, and xAI likely do similarly, with multi-agent systems touted in respective model and product releases~\citep{anthropic2025multiagent,anthropic2026agentteams,openai2026codexsubagents,xai2026grokmultiagent,xai2025grok4,meta2026musespark,google2025deepthink,gottweis2026coscientist}.

\section{Methods}
\label{methods}

We deliberately use the minimal tool definitions necessary to train agentic context management. However, our harness is sufficient to allow arbitrarily complex multi-agent decompositions. Decompositions in the literature fall into roughly three categories: \emph{tree reduce}~\citep{wu2021recursive,sarthi2024raptor,zhang2022summn}, \emph{sequential folds}~\citep{yu2025memagent,zhang2024coa}, and \emph{agentic delegation}~\citep{zhang2025rlm,kimiteam2026kimik25,anthropic2025multiagent}. We utilize all three strategies.

Without any training, models generally make poor use of our harness. Base Qwen3.6-35B-A3B, GPT-5.4, and Kimi K2.6 use subagent delegation sporadically, sometimes choosing to read the entire document in one context, even if context limit is specified to be less than document length, and other times delegating the exact original question to subagents. Opus 4.6 does slightly better, showing real context pressure awareness and document range subagent delegation. Unfortunately it delegates \emph{too} much, blowing through \$300 of API credits building massive subagent trees on a 100-question eval, only to overcount\footnote{Opus 4.6 does appear to perform somewhat decently on our harness with unlimited context budget. Unfortunately, due to the massive subagent trees, running the full comparison without restriction is cost-prohibitive.}. Instead, we compare to base models supplied the entire document in context with native context limits, where performance is stronger.

In general, models have a \emph{weak prior} for context decomposition. Pre-training is pure language continuation, and agentic post-training does not teach decomposition strategies. Where Recursive Language Models~\citep{zhang2025rlm} can rely on strong coding priors instilled in both pre- and post-training, decomposition on our harness needs to be taught.

One idea might be to RL an open weight model on our harness. Unfortunately, exploration in LLM RL is far too weak to escape the weak decomposition prior at modest compute budgets. It is also unclear what reward to assign subagents, and if GRPO~\citep{shao2024deepseekmath} or its derivatives~\citep{yu2025dapo,liu2025drgrpo,zheng2025gspo} are used, what group to compare subagent trajectories in.

We instead SFT a decomposition strategy. We train \emph{tree reduce} for order-independent tasks (see Figure~\ref{fig:treereduce}), \emph{sequential fold} for order-dependent tasks (see Figure~\ref{fig:fold}), and \emph{agentic delegation} to handle boundary conditions whenever we read the document (see Figure~\ref{fig:boundary}).
By not hardcoding a decomposition structure in our harness, our model \emph{internalizes} and \emph{composes} delegation strategies according to the task and current problem-solving state.

\tikzset{
  agent/.style={draw=black!70, rounded corners=3pt, line width=0.6pt, minimum height=6.2mm,
                minimum width=11mm, align=center, font=\footnotesize, inner sep=2pt},
  root/.style={agent, fill=green!22},
  mid/.style={agent, fill=yellow!35},
  leaf/.style={agent, fill=blue!12},
  spawn/.style={-{Stealth[length=2.2mm]}, line width=0.55pt, black!65},
  ret/.style={-{Stealth[length=2.2mm]}, line width=0.55pt, black!65, densely dashed},
  state/.style={font=\scriptsize, text=black, inner sep=1pt},
  note/.style={font=\scriptsize, text=black},
  brace/.style={decorate, decoration={brace, amplitude=4pt}, line width=0.6pt, black!65},
  txt/.style={draw=black!45, rounded corners=2pt, line width=0.4pt, font=\scriptsize, align=left,
              inner sep=1.2mm, text width=\dimexpr\linewidth-2.4mm-1pt\relax},
  qtxt/.style={txt, fill=black!5},
}
\newcommand{\rng}[1]{\ensuremath{[#1)}}
\newcommand{\who}[1]{\textbf{#1}\enspace}
\newcommand{\tool}[1]{\texttt{#1}}
\newcommand{\nsz}{\fontsize{8}{8.5}\selectfont}
\newcommand{\recs}[1]{\begin{tabular}{@{}l@{\enspace}l@{=}r@{\enspace$\to$\enspace}r@{}l@{}r@{}l@{}}#1\end{tabular}}
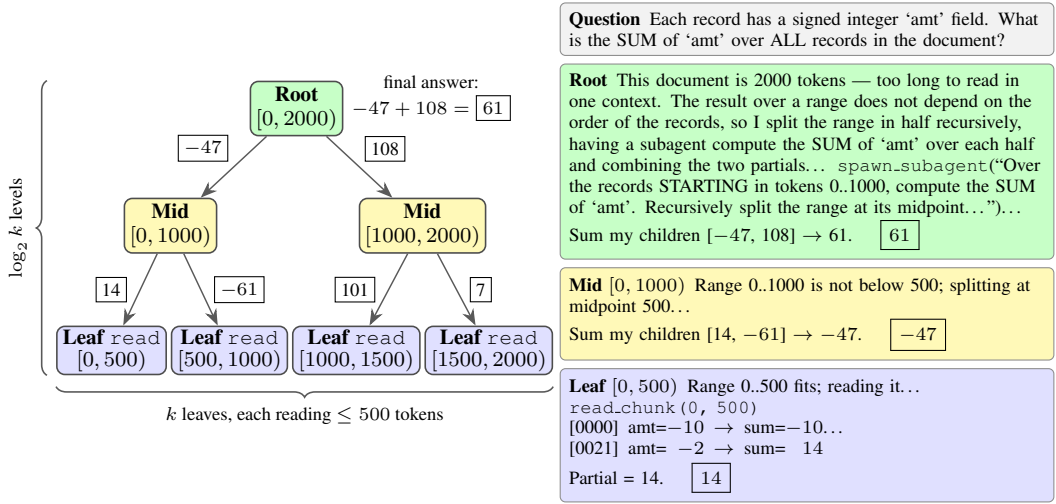
\begin{figure}[H]
\centering
\begin{minipage}[c]{0.525\linewidth}
\begin{tikzpicture}[transform shape]
  \node[leaf, inner sep=1.3pt, inner xsep=2.2pt] (l1) at (0,0) {\nsz\textbf{Leaf}\hspace{2pt}\tool{read}\\[-2pt]\nsz\rng{0,500}};
  \node[leaf, inner sep=1.3pt, inner xsep=2.2pt, right=0.5mm of l1] (l2) {\nsz\textbf{Leaf}\hspace{2pt}\tool{read}\\[-2pt]\nsz\rng{500,1000}};
  \node[leaf, inner sep=1.3pt, inner xsep=2.2pt, right=0.5mm of l2] (l3) {\nsz\textbf{Leaf}\hspace{2pt}\tool{read}\\[-2pt]\nsz\rng{1000,1500}};
  \node[leaf, inner sep=1.3pt, inner xsep=2.2pt, right=0.5mm of l3] (l4) {\nsz\textbf{Leaf}\hspace{2pt}\tool{read}\\[-2pt]\nsz\rng{1500,2000}};
  \node[mid] (m1) at ($($(l1)!0.5!(l2)$)+(0,16.5mm)$) {\nsz\textbf{Mid}\\[-1pt]\nsz\rng{0,1000}};
  \node[mid] (m2) at ($($(l3)!0.5!(l4)$)+(0,16.5mm)$) {\nsz\textbf{Mid}\\[-1pt]\nsz\rng{1000,2000}};
  \node[root] (r) at ($($(m1)!0.5!(m2)$)+(0,15.5mm)$) {\nsz\textbf{Root}\\[-1pt]\nsz\rng{0,2000}};
  \draw[spawn] (r) -- node[state, draw=black, text=black, font=\scriptsize, line width=0.4pt, inner sep=2.2pt, fill=white, above left=2pt and 2pt]  {$-47$} (m1);
  \draw[spawn] (r) -- node[state, draw=black, text=black, font=\scriptsize, line width=0.4pt, inner sep=2.2pt, fill=white, above right=2pt and 2pt] {108}   (m2);
  \draw[spawn] (m1) -- node[state, draw=black, text=black, font=\scriptsize, line width=0.4pt, inner sep=2.2pt, fill=white, left=5pt]  {14}    (l1);
  \draw[spawn] (m1) -- node[state, draw=black, text=black, font=\scriptsize, line width=0.4pt, inner sep=2.2pt, fill=white, right=5pt] {$-61$} (l2);
  \draw[spawn] (m2) -- node[state, draw=black, text=black, font=\scriptsize, line width=0.4pt, inner sep=2.2pt, fill=white, left=5pt]  {101}   (l3);
  \draw[spawn] (m2) -- node[state, draw=black, text=black, font=\scriptsize, line width=0.4pt, inner sep=2.2pt, fill=white, right=5pt] {7}     (l4);
  \node[state, text=black, align=center, right=2pt of r, yshift=4pt] {final answer:\\[2pt]$-47+108={\setlength{\fboxsep}{2.2pt}\fbox{\ensuremath{61}}}$};
  \draw[brace] ([xshift=-1.2mm]l1.south west) -- ([xshift=-1.2mm]l1.west |- r.north)
     node[midway, left=4pt, note, rotate=90, anchor=south] {$\log_2 k$ levels};
  \draw[brace] ([yshift=-1.5mm]l4.south east) -- ([yshift=-1.5mm]l1.south west)
     node[midway, below=4pt, note] {$k$ leaves, each reading $\leq500$ tokens};
\end{tikzpicture}
\end{minipage}\hfill
\begin{minipage}[c]{0.47\linewidth}
\begin{tikzpicture}[node distance=1mm]
  \node[qtxt] (q) {\who{Question}Each record has a signed integer `amt' field. What is the SUM of `amt' over ALL
    records in the document?};
  \node[txt, fill=green!22, below=of q] (t1) {\who{Root}This document is 2000 tokens --- too long to read in one
    context. The result over a range does not depend on the order of the records, so I split the range in half
    recursively, having a subagent compute the SUM of `amt' over each half and combining the two partials\dots
    \tool{ spawn\_subagent}(``Over the records STARTING in tokens 0..1000, compute the SUM of `amt'. Recursively split
    the range at its midpoint\dots'')\dots\\
    Sum my children [$-47$, 108] $\to$ 61. \quad $\boxed{61}$};
  \node[txt, fill=yellow!35, below=of t1] (t2) {\who{Mid \rng{0,1000}}Range 0..1000 is not below 500; splitting at
    midpoint 500\dots\\
    Sum my children [14, $-61$] $\to$ $-47$. \quad $\boxed{-47}$};
  \node[txt, fill=blue!12, below=of t2] (t3) {\who{Leaf \rng{0,500}}Range 0..500 fits; reading it\dots
     \tool{ read\_chunk(0, 500)}\\
    \recs{{[0000]} & amt & $-10$ & & sum= & $-10$ & \dots\\ {[0021]} & amt & $-2$ & & sum= & $14$ &}\\
    Partial = 14. \quad $\boxed{14}$};
\end{tikzpicture}
\end{minipage}

\caption{\textbf{Tree reduce}, for order-independent tasks (aggregation, retrieval). Non-leaf nodes delegate half their range to subagents. Leaves read one chunk and compute a leaf op which is aggregated up the tree.}
\label{fig:treereduce}
\end{figure}

\begin{figure}[H]
\centering
\begin{minipage}[c]{0.378\linewidth}\centering
\begin{tikzpicture}[node distance=6mm]
  \node[root, minimum width=27mm, inner xsep=4pt] (a1) {\nsz\textbf{Root}\enspace\nsz\tool{read} \rng{0,400}};
  \node[mid,  minimum width=27mm, inner xsep=4pt, below=of a1] (a2) {\nsz\textbf{Agent 2}\enspace\nsz\tool{read} \rng{400,800}};
  \node[mid,  minimum width=27mm, inner xsep=4pt, below=of a2] (a3) {\nsz\textbf{Agent 3}\enspace\nsz\tool{read} \rng{800,1200}};
  \node[mid,  minimum width=27mm, inner xsep=4pt, below=of a3] (a4) {\nsz\textbf{Agent 4}\enspace\nsz\tool{read} \rng{1200,1600}};
  \node[leaf, minimum width=27mm, inner xsep=4pt, below=of a4] (a5) {\nsz\textbf{Agent 5}\enspace\nsz\tool{read} \rng{1600,2000}};
  \draw[spawn] (a1) -- node[state, right=2pt] {acc $=-2$} (a2);
  \draw[spawn] (a2) -- node[state, right=2pt] {acc $=7$} (a3);
  \draw[spawn] (a3) -- node[state, right=2pt] {acc $=46$} (a4);
  \draw[spawn] (a4) -- node[state, right=2pt] {acc $=9$} (a5);
  \draw[ret] (a5.west) -- ++(-5mm,0) |- node[note, text=black, pos=0.25, left=1pt, align=right] {final\\answer:\\[2pt]${\setlength{\fboxsep}{2.2pt}\fbox{\ensuremath{23}}}$} (a1.west);
\end{tikzpicture}
\end{minipage}\hfill
\begin{minipage}[c]{0.607\linewidth}
\begin{tikzpicture}[node distance=1mm]
  \node[qtxt] (q) {\who{Question}Process the records in order, keeping a running total of `amt'. Whenever a record
    has grp=RST, reset the running total to 0 (that record's amt is NOT added). What is the final running total
    after the last record?\dots};
  \node[txt, fill=green!22, below=of q] (t1) {\who{Root}This document is 2000 tokens --- too long to read in one
    context. This task is order-dependent: a grp=RST record resets the total, and what a record contributes
    depends on the resets before it --- so I process the document from the start with a running accumulator\dots
    \tool{ read\_chunk(0, 400)}\\
    \recs{{[0000]} & amt & $-10$ & & total= & $-10$ & \dots\\ {[0005]} & grp & RST & RESET,\kern0.3em& total= & 0 & \dots\\ {[0017]} & amt & +14 & & total= & $-2$ &}\\
    $\to$ accumulator = $-2$. Delegating the rest 400..2000 with the updated accumulator.
    \tool{ spawn\_subagent}(``Continue the running accumulator\dots\ over the records STARTING in tokens 400..2000.
    accumulator so far = $-2$\dots'')};
  \node[txt, fill=yellow!35, below=of t1] (t2) {\who{Agent 2}Continue the running accumulator\dots over the
    records STARTING in tokens 400..2000. accumulator so far = $-2$\dots
    \tool{ read\_chunk(400, 800)}\\
    \recs{{[0018]} & amt & $-2$ & & total= & $-4$ & \dots\\ {[0020]} & grp & RST & RESET,\kern0.3em& total= & 0 & \dots\\ {[0034]} & amt & +9 & & total= & 7 &}\\
    $\to$ accumulator = 7. Delegating the rest 800..2000 with the updated accumulator.};
  \node[txt, fill=blue!12, below=of t2] (t3) {\who{Agent 5}\dots accumulator so far = 9\dots Range 1600..2000 fits
    and reaches the document end; reading it.
    \tool{ read\_chunk(1600, 2000)}\\
    \recs{{[0070]} & amt & +18 & & total= & 27 & \dots\\ {[0086]} & amt & $-17$ & & total= & 23 &}\\
    $\to$ accumulator = 23 (final). \quad $\boxed{23}$};
\end{tikzpicture}
\end{minipage}

\caption{\textbf{Sequential fold}, for order-dependent tasks (variable tracking, running state). Each agent reads the next
slice, updates the accumulator, and delegates the rest of the document to the next agent.}
\label{fig:fold}
\end{figure}
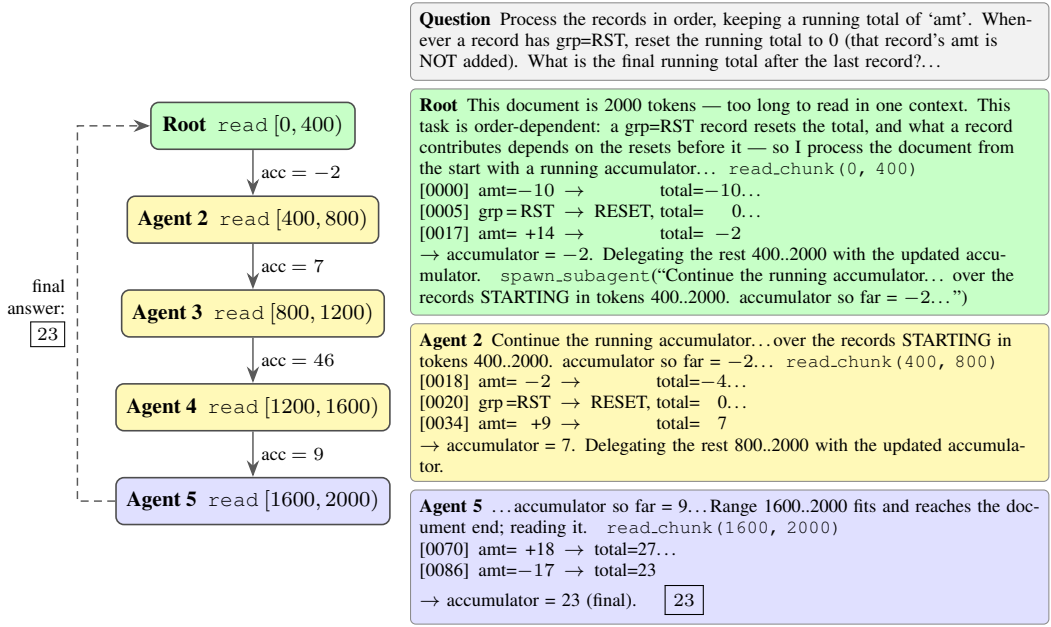

\begin{figure}[H]
\centering
\begin{minipage}[c]{0.32\linewidth}\centering
\begin{tikzpicture}[node distance=7mm]
  \node[agent, fill=blue!12, minimum width=27mm, inner xsep=4pt] (r1) {\nsz\textbf{Leaf}\enspace\nsz\tool{read} \rng{0,500}};
  \node[agent, fill=orange!25, minimum width=27mm, inner xsep=4pt, below=of r1] (r2) {\nsz\textbf{Leaf}\enspace\nsz\tool{read} \rng{500,700}};
  \node[agent, fill=teal!18, minimum width=27mm, inner xsep=4pt, below=of r2] (r3) {\nsz\textbf{Leaf}\enspace\nsz\tool{read} \rng{700,900}};
  \draw[spawn] (r1) -- (r2);
  \draw[spawn, densely dashed] (r2) -- node[note, right=2pt, font=\scriptsize\itshape] {repeat as needed} (r3);
  \node[state, text=black, below=4pt of r3] (ans) {${\setlength{\fboxsep}{2.2pt}\fbox{\ensuremath{2}}}$};
  \draw[ret] (ans.west) -- node[below=1pt, note, text=black] {returned to caller} ++(-22mm,0) -- ++(0,5mm);
\end{tikzpicture}
\end{minipage}\hfill
\begin{minipage}[c]{0.66\linewidth}
\begin{tikzpicture}[node distance=1mm]
  \node[qtxt] (q) {\who{Question}How many entries contain the word `good' anywhere in their body text?};
  \node[txt, fill=blue!12, below=of q] (t1) {\who{Leaf \rng{0,500}}Range 0..500 fits; reading it, then the next 200
    tokens to finish the last entry's body (it may run well past 500).\\
    \tool{read\_chunk(0, 500)} $\to$ {\color{black!55}\itshape Entry: 1\dots\ Entry: 2\dots\ the pity of the
    \textbf{good} God?''\dots\ Entry: 3\dots\ his are not}\\
    \tool{read\_chunk(500, 700)} $\to$ {\color{black!55}\itshape amenable to mere strength.\dots\ ``Now,'' he said, ``friend}};
  \node[txt, fill=orange!25, below=of t1] (t2) {The last entry's body is still cut off at 700 (no blank line / next
    header yet), so I read the next 200 tokens (700..900) to finish it.\\
    \tool{read\_chunk(700, 900)} $\to$ {\color{black!55}\itshape John, where are the skeleton keys?"\dots\ standing him in \textbf{good}
    stead.\dots\ it slowly opened.\enspace Entry: 4\dots}};
  \node[txt, fill=teal!18, below=of t2] (t3) {Computing the partial over the entries whose HEADER STARTS in 0..500
    (\dots any entry whose header starts at/after 500 belongs to the next range):\\
    \begin{tabular}{@{}l@{\enspace}l@{\enspace}l@{\enspace$\to$\enspace}l@{}}
      - entry 1: & `good' appears 0x in the body & (no)     & count=0\\
      - entry 2: & `good' appears 1x in the body & (counts) & count=1\\
      - entry 3: & `good' appears 1x in the body & (counts) & count=2
    \end{tabular}\\
    Partial = 2. \quad $\boxed{2}$};
\end{tikzpicture}
\end{minipage}

\caption{\textbf{Boundary handling.} A leaf/fold agent owns the data that \emph{starts} in its range. It reads 200 additional tokens at a time until the last item in its range is not cut off (additional calls not pictured in Figures~\ref{fig:treereduce} and~\ref{fig:fold}, for brevity).}
\label{fig:boundary}
\end{figure}
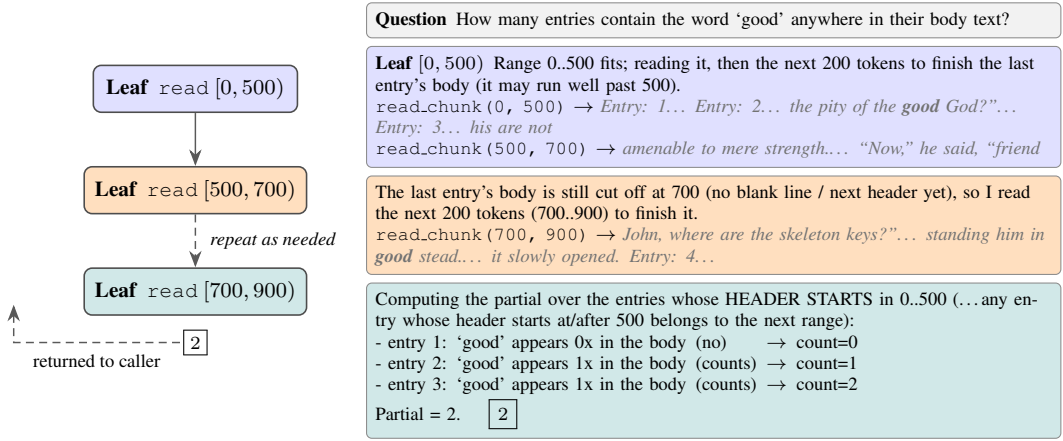

We generate synthetic oracle demonstration data on a wide distribution of aggregation, retrieval, and sequential tasks. While we cover similar shapes to our evals, our training distribution uses independent questions and vastly more diverse examples. The technical challenge is to instill the necessary prior so the model knows how to take a question and delegate appropriate subtasks to agents from which the final answer can be determined.

For example, consider the training question ``In how many sections are there STRICTLY more named than unnamed sentences?" (the prompt defines a \emph{named} sentence as a sentence with a proper noun). To answer this question, each leaf must compute a tally of named and unnamed sentences per section in its assigned range. Thus, the root must recognize the necessary, often multi-dimensional, key space and dictate this and a recursive strategy to mid-level agents, who pass the instruction down to other mid-level agents, until leaves compute their range and counts are aggregated back up the tree, culminating in a final judgement at the root.

Our training data is primarily record-based, but also includes real prose, dot-separated word lists, and haystacks with injected needles. We finetune Qwen3.6-35B-A3B, chosen as a capable small model with affordable compute, trained on the Tinker platform~\citep{thinkingmachines2025tinker,thinkingmachines2025tinkercookbook}. For more details on our complete training data and hyperparameters, see Appendices~\ref{app:data} and~\ref{app:train}, respectively.

We do consider an RL stage following SFT. Unfortunately, even with LLM-as-a-judge rewards on subagent trajectories, as described in Recursive Agent Optimization~\citep{gandhi2026rao}, the policy exhibits high-variance and latches onto simpler strategies that weaken overall performance. We suspect RL to be useful in learning recovery and strengthening generalization; however, to do this properly probably requires group-level baselines on subagent trajectories, to cut variance, and compute beyond the scope of this paper.

\section{Experiments}
\label{experiments}

We test our finetune on RULER~\citep{hsieh2024ruler}, the standard long context eval, and OOLONG-synth~\citep{bertsch2025oolong}, a newer long context benchmark frontier models have not yet saturated, also used in Recursive Language Models~\citep{zhang2025rlm}, Recursive Agent Optimization~\citep{gandhi2026rao}, and Recursive Agent Harnesses~\citep{lumer2026rah}. RULER is an extension of needle-in-a-haystack style tasks with variable tracking, word frequencies, and Q\&A. OOLONG-synth concatenates classification tasks on ten datasets into long documents on which aggregation is queried.

We compare against base Qwen3.6-35B-A3B\footnote{While Qwen3.6-35B-A3B was trained with a 256K token context limit, we sample Qwen3.6-35B-A3B from Tinker, which caps the model at 64K token context.} and GPT-5.4. We chose GPT-5.4 as a selection from the best frontier models available upon Qwen3.6-35B-A3B release. While we do not eval against additional frontier models, due to compute requirements, we recommend the reader read the OOLONG paper for more comparisons (GPT-5 tops the chart)~\citep{bertsch2025oolong}. While we do eval GPT-5.4 on RULER, this benchmark is largely saturated for frontier models. For more RULER reference points, consult model release cards~\citep{nvidia2025nemotron3,glm5team2026glm5,lai2026msa,zhu2026pokeeisaac,dangel2026subq}.

For RULER, we fix a sample of 5 questions per problem type, for a total of 65 problems per document length. For OOLONG-synth, we fix a sample of 10 questions per family: counting, user, and temporal, each sampled from a different underlying classification dataset. We vendor generators for both benchmarks and generate our problems from scratch. We eval our finetune at temperature 0.2 (determined best empirically), GPT-5.4 at API default (required for reasoning mode), and base Qwen3.6-35B-A3B at temperature 0 (we found temperature 0 and 0.2 to perform similarly). We disable thinking on base Qwen3.6-35B-A3B as we found it to improve performance; both modes produce chain-of-thought tokens before boxing an answer.

On RULER, our finetune performs worse than GPT-5.4, while remaining above 85\%. On OOLONG-synth, our model starts worse than GPT-5.4, then matches its performance when document size exceeds 40K tokens. At this length, native context performance begins to degrade, and a single model can be overwhelmed with the number of items requiring tracking within its context. On the other hand, agents in our harness operate at 8K token budgets, executing small pieces of an overall decomposition strategy, which scales gracefully to arbitrary lengths.

Exact scores for Figure~\ref{fig:length-sweep} are given in Table~\ref{tab:comparison}, and per problem-family results for our model in Table~\ref{tab:family-results}.

A quick note on cost: while we do not quantify costs formally, our finetune is cheaper both by running a smaller model and by scaling \emph{agents} over document length instead of context window. In particular, our computational complexity is $O(n)$ in document size $n$, compared to $\Omega(n^2)$ for dense attention architectures.

Our finetune is also \emph{faster} than full document comparisons, particularly when the \emph{tree reduce} decomposition strategy is chosen, which allows for parallel processing.

\begin{table}[H]
\centering\small
\begin{tabular}{@{}lcccccc@{}}
\toprule
RULER (mean of 13 tasks) & 10K & 20K & 40K & 80K & 160K & 320K \\
\midrule
GPT-5.4 (full document)                & \textbf{0.954} & 0.938 & \textbf{0.954} & \textbf{0.954} & \textbf{0.929} & \textbf{0.954} \\
Qwen3.6-35B-A3B (full document)        & 0.923 & 0.938 & 0.938 & ---   & ---   & --- \\
Finetuned Qwen3.6-35B-A3B (our harness)             & 0.949 & \textbf{0.949} & 0.929 & 0.885 & 0.868 & 0.858 \\
\midrule
OOLONG-synth (mean of 3 families) & 10K & 20K & 40K & 80K & 160K & 320K \\
\midrule
GPT-5.4 (full document)                & \textbf{0.718} & \textbf{0.666} & \textbf{0.600} & 0.539 & \textbf{0.556} & \textbf{0.479} \\
Qwen3.6-35B-A3B (full document)        & 0.536 & 0.558 & 0.388 & ---   & ---   & --- \\
Finetuned Qwen3.6-35B-A3B (our harness)             & 0.606 & 0.540 & 0.535 & \textbf{0.561} & 0.464 & 0.470 \\
Random baseline                        & 0.226 & 0.225 & 0.184 & 0.181 & 0.191 & 0.161 \\
\bottomrule
\end{tabular}
\caption{\textbf{Results by document length} (tabular view of Figure~\ref{fig:length-sweep}). Our finetune runs in our harness with an 8K per-agent context limit. GPT-5.4 (medium reasoning, 1M token context limit) and base Qwen3.6-35B-A3B (reasoning disabled, 64K token Tinker serving limit) receive the full document in one call (see Appendix~\ref{app:harness} for details). Bold marks the highest score per length.}
\label{tab:comparison}
\end{table}

\begin{table}[H]
\centering\footnotesize\setlength{\tabcolsep}{3.5pt}
\begin{minipage}[c]{0.535\linewidth}\centering
\begin{tabular}{@{}l@{\hspace{3.5pt}}cccccc@{}}
\toprule
RULER & 10K & 20K & 40K & 80K & 160K & 320K \\
\midrule
NIAH (8) & 1.00 & 1.00 & 1.00 & 1.00 & 1.00 & 1.00 \\
VT             & 1.00 & 1.00 & 1.00 & 0.92 & 0.84 & 0.64 \\
CWE            & 1.00 & 0.94 & 0.74 & 0.58 & 0.24 & 0.18 \\
FWE            & 0.93 & 1.00 & 0.93 & 1.00 & 1.00 & 0.93 \\
QA-1           & 0.60 & 0.60 & 0.80 & 0.60 & 0.60 & 0.60 \\
QA-2           & 0.80 & 0.80 & 0.60 & 0.40 & 0.60 & 0.80 \\
\midrule
Mean (13) & 0.949 & 0.949 & 0.929 & 0.885 & 0.868 & 0.858 \\
\bottomrule
\end{tabular}
\end{minipage}\hfill
\begin{minipage}[c]{0.455\linewidth}\centering
\begin{tabular}{@{}l@{\hspace{3.5pt}}c@{\hspace{3.5pt}}c@{\hspace{3.5pt}}c@{\hspace{3.5pt}}c@{\hspace{3.5pt}}c@{\hspace{3.5pt}}c@{}}
\toprule
OOLONG & 10K & 20K & 40K & 80K & 160K & 320K \\
\midrule
Counting & 0.45 & 0.42 & 0.42 & 0.41 & 0.41 & 0.40 \\
User     & 0.86 & 0.72 & 0.80 & 0.73 & 0.61 & 0.60 \\
Temporal & 0.51 & 0.48 & 0.39 & 0.54 & 0.37 & 0.41 \\
\midrule
Mean     & 0.606 & 0.540 & 0.535 & 0.561 & 0.464 & 0.470 \\
\bottomrule
\end{tabular}
\end{minipage}
\caption{\textbf{Results by task family}. \textbf{Left:} On RULER, our model chooses a lossy aggregation strategy for the common words ``CWE" problem type, which degrades when common words are diluted by fillers as document length grows. \textbf{Right:} On OOLONG-synth, leaf classification and merge errors compound at length. Distribution mismatch also arises at 160K+ document length (it sees $\leq $14K doc size in training).}
\label{tab:family-results}
\end{table}

\section{Conclusion}
\label{conclude}

In this paper we show training on a bare bones agentic harness is sufficient for a small model with small context to perform as well as a frontier model with 1M token context on long-context evals. While we do not claim full generalization of our finetune, we also do not train on our evaluation distributions. The point of this paper is to show the power of agentic training in contrast to harness engineering, excessive native long context, and lossy compaction. As a side effect we also show the power of synthetic oracle SFT data generated by frontier models (Opus 4.8, 5, 5.5, and Fable 5 and 5.1, in this case).

The thesis of the last decade of AI development is that specific model capability can be incorporated into fully-general models by training on a wider distribution. The methods of this paper can be incorporated into frontier model development, so such models exhibit improved agentic performance under context pressure, without wasting tokens. This is in contrast to an external harness which must be applied at runtime.

To develop our curriculum into a production recipe, the training distribution would need to be widened and a reinforcement learning stage would likely prove fruitful in supercharging agentic capabilities. We leave this as important future work, outside the scope of this paper or a small research project more generally.

\subsection*{AI use statement}

In this work, we used generative AI tools to write all our code, administer experiments, create figures, search and analyze relevant literature, format references, generate synthetic datasets, propose or refine hypotheses/brainstorming, design or provide feedback on research methodology/experiments, clean and reformat data, support qualitative and thematic data analysis, and interpret results. We have not used generative AI tools to draft parts of this research paper, and the following topics are not applicable to this work: help develop theoretical models or conceptual frameworks, formulate mathematical claims, provide critical ingredients for proving mathematical claims, assist in the writing of proofs, and assist with translation. We have reviewed our AI-assisted work, for example, by reviewing AI code and conducting manual audits of synthetic data. We take responsibility for the final content of this work,
including text, claims or artifacts produced with the aid of generative AI.

\raggedbottom
\bibliography{iclr2027_conference}
\bibliographystyle{iclr2027_conference}

\pagebreak

\clearpage
\appendix
\section{Harness and prompts}
\label{app:harness}

The full system prompt and tool descriptions are given in Figure~\ref{fig:system-prompt}. The full-document prompt, used for single shot baselines, is given in Figure~\ref{fig:full-doc-prompt}. Benchmark text fields are supplied in Table~\ref{tab:prompt-slots}. Harness responses are given in Table~\ref{tab:tool-results}.

\newcommand{\slot}[1]{\textrm{\textit{$\langle$#1$\rangle$}}}
\newcommand{\promptbox}[1]{\noindent\fbox{\parbox{\dimexpr\linewidth-2\fboxsep-2\fboxrule\relax}{\footnotesize\ttfamily
  \raggedright\setlength{\parskip}{0.6em}#1}}}

\begin{figure}[H]
\promptbox{%
You are a long-document assistant. The document is \slot{N} tokens long; you cannot see it directly. Your own
context window is \slot{B} tokens {\rmfamily\textemdash} the conversation (system prompt, user message, your responses, and all
tool results) must fit in this budget or the episode ends.

\slot{task context, if the benchmark provides one}

You have two tools:\\
- `read\_chunk(start, end)`: read the document tokens in [start, end).\\
- `spawn\_subagent(subtask)`: delegate to a fresh-context copy of yourself (also \slot{B} tokens) with `subtask` as
the user prompt. The same tools and system prompt as above are inherited by the subagent.

When you are confident in the final answer, emit it as \textbackslash{}boxed\{value\} and stop.}

\vspace{2mm}
\promptbox{%
\textrm{\textbf{Tool descriptions}} (function-calling schema)\\[0.3em]
read\_chunk: Read a slice of the document and return the decoded text of tokens [start, end). The document has a
fixed length (stated in the system prompt). Each call is capped at the chunk limit; for larger ranges, issue multiple
reads or delegate to a subagent.\\
\quad start: First token position to read (inclusive).\enspace end: Last token position to read (exclusive).\\[0.3em]
spawn\_subagent: Spawn a fresh-context copy of yourself to solve `subtask`; returns the child's \textbackslash{}boxed\{\}
answer.\\
\quad subtask: Sub-problem statement (free text) for the child to solve}
\caption{\textbf{Agent system prompt}, shared by the root and every subagent, and the two tool descriptions. \slot{N}
is the document length and \slot{B} the per-agent context budget. The root's user message is the benchmark question
(Table~\ref{tab:prompt-slots}); a subagent's user message is the \texttt{subtask} string its parent wrote.}
\label{fig:system-prompt}
\end{figure}

\begin{figure}[H]
\promptbox{%
\textrm{\textbf{System}}\\[0.3em]
\slot{task context, if the benchmark provides one}

You are given a complete document followed by a question about it. Read the document carefully and answer the
question. Show your reasoning, then emit your final answer as \textbackslash{}boxed\{value\} and stop.}

\vspace{2mm}
\promptbox{%
\textrm{\textbf{User}}\\[0.3em]
\slot{full document text}

\slot{question}}
\caption{\textbf{Full-document prompt}, used for the single-shot baselines: the whole document is placed in one
user message, with no tools and no budget prose. For RULER, whose template embeds the document mid-prompt, the
document is instead placed back at the template's \texttt{\{context\}} position, reproducing RULER's own prompt.}
\label{fig:full-doc-prompt}
\end{figure}

\begin{table}[H]
\centering\small
\begin{tabular}{@{}p{0.15\linewidth}p{0.37\linewidth}p{0.4\linewidth}@{}}
\toprule
Benchmark & System prompt task context & Root user message \\
\midrule
OOLONG-synth & OOLONG's dataset description, verbatim (its intro, the number of items, the label set) &
  OOLONG's question, verbatim, + ``Put your final answer inside \textbackslash{}boxed\{\}.'' \\[0.4em]
RULER & None (empty) & RULER's task template, verbatim, with \texttt{\{context\}} replaced by a pointer to the
  document (below) + ``Reply with the answer inside \textbackslash{}boxed\{\}.'' \\
\bottomrule
\end{tabular}
\caption{\textbf{Benchmark text fields.} The task context is placed in the system prompt, so every
subagent inherits it without the parent restating it. For RULER, \texttt{\{context\}} in the template is replaced
by: ``[The relevant text is in a separate document accessible via the read\_chunk tool --- see the system prompt for
usage.] (Document length: \slot{N} tokens.)''}
\label{tab:prompt-slots}
\end{table}

\begin{table}[!ht]
\centering\small
\begin{tabular}{@{}p{0.42\linewidth}p{0.53\linewidth}@{}}
\toprule
Event & Tool result returned to the calling agent (verbatim) \\
\midrule
\multicolumn{2}{@{}l}{\textit{\texttt{spawn\_subagent}}} \\
\cmidrule{1-2}
Subagent ends with a boxed answer & The content of its last closed \texttt{\textbackslash boxed\{\ldots\}} \\
Subagent exceeds its context budget & \texttt{agent overflowed context} \\
Subagent stops without a boxed answer & \texttt{agent did not box an answer} \\
Tree node cap reached & \texttt{Error: rollout node budget reached. Solve directly.} \\
\midrule
\multicolumn{2}{@{}l}{\textit{\texttt{read\_chunk}}} \\
\cmidrule{1-2}
Valid range & The decoded text of tokens \texttt{[start, end)}, clamped to the document \\
Empty range after clamping & \texttt{Empty range.} \\
Non-integer arguments & \texttt{Error: read\_chunk requires integer `start' and `end'.} \\
\bottomrule
\end{tabular}
\caption{\textbf{Harness responses.} An agent's budget counts its whole rendered conversation (system
prompt, tool schema, user message, its own turns and every tool result); each generation is capped at the remaining
room. A subagent that answers returns only its boxed
content, so its reads and reasoning never enter the parent's context.}
\label{tab:tool-results}
\end{table}

\clearpage
\section{Training data}
\label{app:data}

A high-level description of our training tasks is given in Table~\ref{tab:corpus}. 

\begin{table}[!ht]
\centering\footnotesize
\begin{tabular}{@{}l@{\hspace{5pt}}l@{\hspace{5pt}}r@{\hspace{7pt}}p{0.52\linewidth}@{}}
\toprule
Task & Strategy & $N$ & Example question / what the leaf does \\
\midrule
\multicolumn{4}{@{}l}{\textit{Order-independent aggregation over structured records} (one record per line)} \\
\cmidrule{1-4}
\texttt{synth\_sum} & tree & 60 & SUM of \texttt{amt} over all records \\
\texttt{synth\_count} & tree & 60 & how many records have \texttt{flag=Y} \\
\texttt{synth\_max} & tree & 60 & largest \texttt{amt} \\
\texttt{synth\_min} & tree & 60 & smallest \texttt{amt} \\
\texttt{synth\_sumwhere} & tree & 60 & SUM of \texttt{amt} over only the \texttt{flag=Y} records \\
\texttt{synth\_maxwhere} & tree & 60 & largest \texttt{amt} among a flag-filtered subset \\
\texttt{synth\_mode} & tree & 60 & most common \texttt{grp} (per-\texttt{grp} tally) \\
\texttt{synth\_distinct} & tree & 60 & how many distinct \texttt{grp} values (set state) \\
\texttt{synth\_sumby} & tree & 60 & which \texttt{grp} has the largest total \texttt{amt} \\
\texttt{synth\_count2} & tree & 60 & count under a conjunction of two field conditions \\
\texttt{synth\_diff} & tree & 60 & SUM over \texttt{flag=Y} minus SUM over \texttt{flag=N} \\
\texttt{synth\_count\_cmp} & tree & 60 & how many records have \texttt{amt} below a threshold \\
\texttt{synth\_count\_range} & tree & 60 & how many records have \texttt{amt} in an interval \\
\texttt{synth\_filter\_argmax} & tree & 90 & most / least common value among a filtered subset \\
\texttt{synth\_2d} & tree & 128 & two-level questions (e.g.\ for how many months is \texttt{K3} the most common group) \\
\midrule
\multicolumn{4}{@{}l}{\textit{Sequential folds}} \\
\cmidrule{1-4}
\texttt{synth\_runreset} & fold & 70 & running total that resets at \texttt{grp=RST} \\
\texttt{synth\_varchain} & fold & 70 & final value of a variable under \texttt{set B = A} copies \\
\texttt{synth\_peak} & fold & 70 & highest value the running total reaches \\
\texttt{synth\_streak} & fold & 70 & longest run of consecutive \texttt{flag=Y} records \\
\texttt{synth\_adjacent} & fold & 70 & records whose \texttt{amt} exceeds the previous record's \\
\texttt{synth\_first\_exceed} & fold & 70 & index where the running total first exceeds a threshold \\
\texttt{vt\_novel} & fold & 600 & variable-assignment chains hidden in prose \\
\midrule
\multicolumn{4}{@{}l}{\textit{Open-vocabulary frequency}} \\
\cmidrule{1-4}
\texttt{synth\_topk} & tree & 200 & the $k$ most frequent words in a word list; partials are pruned \texttt{word:count} tallies \\
\midrule
\multicolumn{4}{@{}l}{\textit{Multi-line records}} \\
\cmidrule{1-4}
\texttt{long\_records} & tree & 400 & header + prose-body entries; counts, argmaxes, 2-D tallies, and word mentions in bodies \\
\midrule
\multicolumn{4}{@{}l}{\textit{Labelling real text, then aggregating}} \\
\cmidrule{1-4}
\texttt{labeled\_records} & tree & 880 & items from 7 labelled datasets; the leaf judges each label, then counts / compares \\
\texttt{rule\_label} & tree & 240 & prose sentences labelled by a stated rule (e.g.\ ends with ``?''), then aggregated \\
\texttt{realdoc\_count} & tree & 240 & occurrences of a word in a Gutenberg novel passage \\
\midrule
\multicolumn{4}{@{}l}{\textit{Retrieval and QA in prose}} \\
\cmidrule{1-4}
\texttt{niah\_novel} & tree & 200 & find the one sentence stating a key's magic value \\
\texttt{niah\_multi} & tree & 220 & collect several hidden facts, or a lookup instruction plus all facts, and resolve at the root \\
\texttt{niah\_bridge} & tree & 80 & two-hop: fact 1 names who holds a role, fact 2 answers about that person \\
\texttt{narrativeqa} & tree & 140 & short-answer questions over book / script windows; model-executed leaf \\
\midrule
Total & & 4{,}618 & \\
\bottomrule
\end{tabular}
\caption{\textbf{Training tasks.} $N$ is the number of problems per task. \emph{Tree}: recursive halving, with
each parent merging its two children's partials (for retrieval, the half that finds the answer wins). \emph{Fold}: one
agent reads a slice and hands the rest of the document and its accumulator to the next. Every agent in each tree is a
training example.}
\label{tab:corpus}
\end{table}

We upweight root gradient mass 4x and positive \texttt{narrativeqa} leaves 2x, downsampling negative leaves by half. Each task has prompt jitter to train a wider distribution. We also jitter the system prompt. We train with context budgets 3K (50\%), 5K (15\%), 8K (13\%), 10K (11\%), and 12K (11\%), and document length between 4K and 14K tokens, with a distribution so document length is typically larger than context budget.

\section{Training details}
\label{app:train}

Our training runs used rank 32 LoRA on Qwen3.6-35B-A3B, chat renderer \verb|qwen3_5_disable_thinking|, cross-entropy loss on assistant tokens only. Small batch training of size 8-64 (you pay per token on Tinker, and small batch training is generally better-behaved per token, particularly on LoRAs~\citep{marek2025smallbatch,schulman2025lora,masters2018small,keskar2017largebatch}). We used pure Adam optimizer, no weight decay, no gradient clipping, $\eps=$1e-12, $\beta_1 = 0.9$, $\beta_2 = 0.95$, with constant learning rate between 5e-6 and 1e-5, single epoch training.

\end{document}